\documentclass[11pt]{article}

\usepackage[final]{acl}

\usepackage{times}
\usepackage{latexsym}
\usepackage{booktabs}
\usepackage{hyperref}
\usepackage[T1]{fontenc}

\usepackage[utf8]{inputenc}

\usepackage{microtype}

\usepackage{inconsolata}

\usepackage{graphicx}
\usepackage{xcolor}

\title{VeriDx: Earning the Right to Diagnose with Disease-Centric Verification}

\author{
  \textbf{Zhong Cao}\thanks{\raggedright Correspondence: \href{mailto:zhong.cao@uni-heidelberg.de} {\texttt{zhong.cao@uni-heidelberg.de}}.} \\
  Heidelberg University
  \And
  \textbf{Shuying Chen} \\
  University of International Business and Economics
}

\begin{document}
\maketitle

\begin{abstract}
A correct diagnosis can still be reached for the wrong reasons. In clinical reasoning, every disease hypothesis creates obligations: key evidence must be checked, alternatives must be ruled out, contradictions must be resolved, useful tests must be considered, and closure must be justified. Current evaluations of medical LLMs mostly focus on final answers, local steps, or isolated facts, and therefore miss these hypothesis-induced commitments. We introduce \textbf{VeriDx}, a disease-centric verification framework that links free-form diagnostic reasoning to structured disease profiles. VeriDx tracks whether each hypothesis is satisfied, unresolved, or violated its clinical obligations, exposing failures such as missing critical tests, unresolved differentials, ignored contradictions, unsupported claims, and premature closure. We instantiate VeriDx for complex respiratory diagnosis using guideline-derived disease profiles and expert-annotated longitudinal cases. Our results show that many diagnostic errors are not isolated mistakes, but broken commitments made earlier in the reasoning process.
\end{abstract}

\section{Introduction}

\begin{figure*}[t]
  \includegraphics[width=\textwidth]{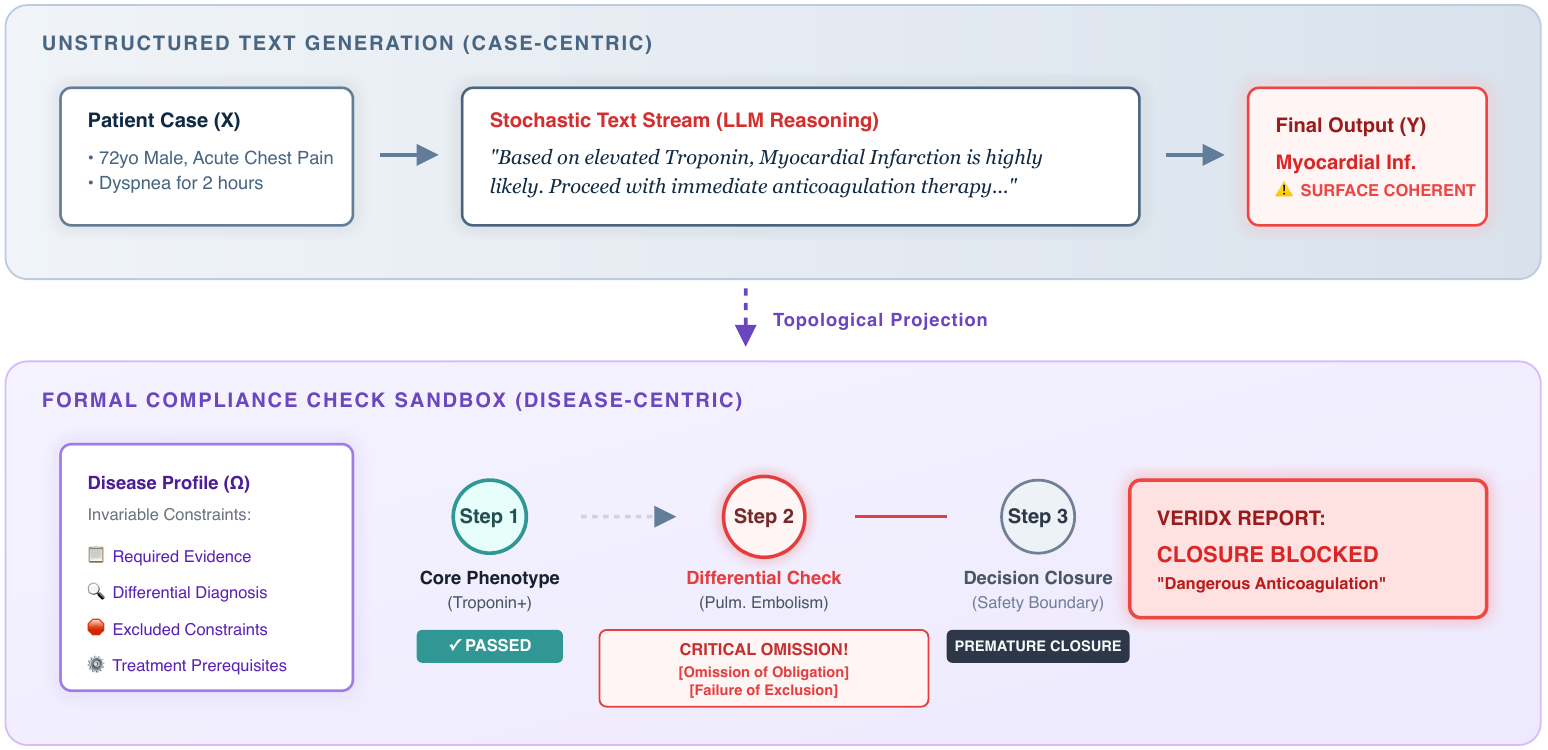}
  \caption{Disease-centric DDx profiles for standardized diagnostic verification. Conventional case-centric LLM reasoning can produce fluent and locally plausible clinical narratives while inconsistently covering disease-specific diagnostic obligations. VeriDx introduces disease-centric DDx profiles that encode required evidence, differential diagnoses, exclusion constraints, and treatment prerequisites for each disease. These profiles can be automatically updated from clinical guidelines and evidence, and are used to guide and audit LLM-generated reasoning. By mapping free-text diagnostic narratives onto profile-defined obligations, VeriDx enables consistent, auditable, and safety-aware verification across cases.}
  \label{fig:illstation}
\end{figure*}

Clinical diagnosis should be verified not only by the final disease label, but also by whether the reasoning process satisfies the obligations created by each disease hypothesis. Once a model proposes pulmonary embolism, it should address embolism exclusion, imaging evidence, risk stratification, and alternative causes of dyspnea; once it proposes interstitial lung disease, it should examine autoimmune disease, infection, exposure history, imaging patterns, and longitudinal progression. These obligations define whether a diagnostic conclusion is clinically responsible. We therefore introduce \textbf{VeriDx}, a disease-centric verification framework that makes such hypothesis-induced obligations explicit and checks whether LLM diagnostic reasoning satisfies, resolves, or violates them.

This problem is becoming urgent because medical LLMs are increasingly good at producing plausible clinical answers. Med-PaLM 2 shows strong performance on medical question answering \cite{singhal_toward_2025}, AMIE demonstrates promising diagnostic dialogue ability \cite{tu_towards_2025}, and recent work reports that frontier models can outperform many physicians on difficult diagnostic reasoning tasks \cite{brodeur_performance_2026}. Medical agents are also moving beyond static question answering toward workflow interaction, EHR navigation, tool use, and treatment planning, as evaluated by MedAgentBench and MedAgentBoard \cite{jiang_medagentbench_2025,zhu_medagentboard_2025}. However, stronger answer generation makes reasoning errors harder, not easier, to detect. Recent evaluations show that LLMs still miss critical evidence, select inappropriate examinations, fixate on early hypotheses, overlook contradictions, and close cases prematurely \cite{qiu_quantifying_2025,chiu_simulating_2025}. Thus, the key challenge is no longer only whether a model knows medicine, but whether its reasoning can be audited against clinical obligations.

Existing clinical reasoning benchmarks reveal this challenge, but they do not directly verify hypothesis-induced obligations. MedR-Bench \cite{qiu_quantifying_2025}, DiagnosisArena \cite{zhu_diagnosisarena_2026}, VivaBench \cite{chiu_simulating_2025}, and MedAgentBench \cite{jiang_medagentbench_2025} evaluate multi-step diagnosis, examination recommendation, hypothesis refinement, longitudinal interaction, and treatment planning. These benchmarks move beyond final-answer accuracy and expose process-level failures such as missing investigations, weak differentials, and premature closure. Yet their unit of evaluation is still mostly the case, the step, or the next action. They ask whether a system solves a case or performs well at a stage, but not whether each disease hypothesis introduced by the model has fulfilled its own clinical commitments. This gap matters because a model may reach the correct diagnosis while still skipping a discriminative test, leaving a dangerous alternative unresolved, or ignoring evidence that contradicts its preferred hypothesis.

Prior methods for structuring or verifying medical reasoning are useful, but they leave a gap between rigid clinical rules and unconstrained language reasoning. Symbolic and guideline-based systems, including MYCIN \cite{shortliffe_rule-based_1974}, GLIF3 \cite{peleg_glif3_2000}, computable guideline modeling \cite{scott_modelling_2023}, and CPGPrompt \cite{deng_cpgprompt_2026}, show that explicit medical structure can make reasoning more interpretable and auditable. However, real diagnosis involves temporal evidence, negation, subjective symptoms, multimodal findings, incomplete information, and contextual exceptions that are difficult to fully encode as executable rules \cite{deng_cpgprompt_2026}. In parallel, general LLM verification methods such as CoVe \cite{dhuliawala_chain--verification_2024}, Reflexion \cite{shinn_reflexion_2023}, CRITIC \cite{gou_critic_2024}, RARR \cite{gao_rarr_2023}, and self-verification \cite{weng_large_2023} improve factuality, consistency, attribution, or answer selection. Medical consistency-verification methods further improve calibration and reliability in medical question answering \cite{martinez_multi-agent_2026}. However, these methods are mostly domain-agnostic: they can check whether a statement is supported, but they do not specify what must be checked because a particular disease hypothesis was raised.

VeriDx addresses this gap by treating each disease hypothesis as a structured commitment. For every disease, VeriDx builds a disease profile that encodes required evidence, exclusion criteria, differential diagnoses, discriminative examinations, contradiction patterns, treatment-relevant constraints, and closure conditions. Given a free-form reasoning trajectory, VeriDx maps the model's hypotheses, evidence claims, test recommendations, differential updates, contradictions, and closure decisions into a typed reasoning graph linked to these disease profiles. The verifier then tracks whether each obligation is satisfied, unresolved, or violated. In this way, failures such as unsupported claims, ignored contradictions, missing discriminative tests, unresolved differentials, and premature closure become machine-verifiable violations of disease-specific commitments rather than vague defects in natural-language reasoning.

We instantiate VeriDx in complex respiratory diagnosis, where disease-centric obligations are especially important. Respiratory diseases often share symptoms such as cough, dyspnea, fever, hypoxemia, and abnormal chest imaging, so safe diagnosis depends on distinguishing overlapping hypotheses rather than matching a single presentation. Evidence also evolves across time: imaging, laboratory tests, microbiology, exposure history, comorbidities, and treatment response may support different hypotheses at different stages. To evaluate reasoning under these conditions, we construct a respiratory reasoning resource with two components. On the knowledge side, we collect 2,563 respiratory guidelines and reference documents totaling approximately 3.93 GB, from which we normalize diagnostic and therapeutic reasoning constraints for almost all respiratory diseases. On the clinical side, we curate 188 complex real-world respiratory cases with expert annotations, including 1,455 medical images and three-stage longitudinal diagnostic trajectories comprising nine structured sections and averaging 20k--30k tokens per case.


Our work makes three contributions. First, we formulate \emph{disease-centric verifiable reasoning}, which reframes diagnostic reasoning as the satisfaction, verification, and revision of obligations induced by disease hypotheses. Second, we introduce VeriDx, a neuro-symbolic framework that links free-form LLM reasoning trajectories to structured disease profiles and verifies clinically meaningful reasoning failures. Third, we construct a large-scale guideline-derived disease-profile library covering a broad range of respiratory diseases and evaluate VeriDx on 188 expert-annotated longitudinal pulmonary cases, enabling evaluation beyond final-answer accuracy toward auditable, revisable, and disease-grounded clinical reasoning. Together, these contributions shift clinical AI evaluation from asking whether a model gets the answer right to asking whether it has earned the right to diagnose.

\section{Related Work}

\paragraph{Medical LLMs and clinical reasoning evaluation.}
Medical LLMs and clinical agents have advanced rapidly, from medical question answering and diagnostic dialogue to workflow-oriented diagnosis, tool use, and treatment planning \cite{singhal_toward_2025,tu_towards_2025,jiang_medagentbench_2025,zhu_medagentboard_2025}. Alongside this progress, benchmarks such as MedR-Bench, DiagnosisArena, VivaBench, and MedAgentBench evaluate multi-step diagnosis, examination recommendation, hypothesis refinement, and longitudinal interaction, revealing failures such as missed evidence, weak differentials, overlooked contradictions, and premature closure \cite{qiu_quantifying_2025,zhu_diagnosisarena_2026,chiu_simulating_2025,jiang_medagentbench_2025}. VeriDx builds on this shift from answer evaluation to process evaluation, but verifies reasoning through the obligations induced by each disease hypothesis rather than only through case-level success or step-level quality.

\paragraph{Retrieval-augmented and knowledge-grounded reasoning.}
Existing clinical reasoning frameworks improve model performance by retrieving, representing, or organizing external medical knowledge. Retrieval-augmented methods, such as Almanac and Self-BioRAG, retrieve relevant clinical evidence to ground model responses \cite{zakka_almanac_2024,jeong_improving_2024}. Knowledge-guided approaches, such as KARE, incorporate structured resources such as knowledge graphs into healthcare reasoning \cite{jiang_reasoning_2025}, while evidence-organization methods, such as Tree-of-Reasoning and GuideTree, structure clinical evidence or guidelines to support complex diagnostic reasoning \cite{peng_tree_2025,ge_guidetree_2026}. These approaches primarily improve how clinical knowledge is accessed, represented, or organized during diagnosis. However, retrieving relevant knowledge does not guarantee clinically complete reasoning: a model may still fail to distinguish dangerous alternatives, request critical examinations, resolve contradictory findings, or avoid premature diagnostic closure \cite{kim_rethinking_2025}. VeriDx complements these approaches by treating such requirements as explicit, disease-specific clinical obligations and tracking whether each obligation is satisfied, unresolved, or violated throughout the diagnostic process.

\paragraph{Structured clinical knowledge and LLM verification.}
Structured medical reasoning systems, including MYCIN, GLIF3, computable guideline models, and CPGPrompt, show that explicit clinical knowledge can improve interpretability and auditability \cite{shortliffe_rule-based_1974,peleg_glif3_2000,scott_modelling_2023,deng_cpgprompt_2026}, while general LLM verification and repair methods such as CoVe, Reflexion, CRITIC, RARR, and self-verification improve factuality, consistency, attribution, or answer selection \cite{dhuliawala_chain--verification_2024,shinn_reflexion_2023,gou_critic_2024,gao_rarr_2023,weng_large_2023}. Medical consistency-verification methods further improve calibration and reliability in medical QA \cite{martinez_multi-agent_2026}. However, these approaches usually verify statements, answers, or generic reasoning traces, but do not specify what must be checked once a particular disease hypothesis is raised. VeriDx combines flexible LLM reasoning with disease-specific profiles, asking whether the clinical commitments associated with each proposed diagnosis are satisfied, unresolved, or violated.

\section{Data}\label{sec:data}

VeriDx uses two complementary resources for disease-centric verification: a guideline-derived disease-profile library constructed in this work to define diagnostic obligations, and a separately developed longitudinal pulmonary benchmark used to evaluate models under evolving information constraints.

\paragraph{Guideline-derived disease profiles.} We processed 2,563 respiratory guidelines and reference documents ($\sim$3.93 GB) to build structured profiles covering nearly all pulmonary diseases. These profiles serve as the clinical backbone of VeriDx, encoding hypothesis-induced obligations such as required evidence, important mimics, discriminative tests, and closure conditions. Professional physicians reviewed these profiles to ensure the clinical validity and discriminative value of each obligation.


\paragraph{Evaluation benchmark.}
We evaluate VeriDx on a separately developed longitudinal pulmonary benchmark containing 188 complex, de-identified real-world cases. Unlike retrospective summaries that expose the complete diagnostic trajectory, the benchmark preserves the evolving diagnostic process, including incomplete early evidence, examination results, and subsequent diagnostic revisions. Each case is organized into a three-stage longitudinal trajectory comprising nine structured sections, with progressive information disclosure ensuring that the model can access only the evidence available at the current clinical stage. The resulting evaluation contains 1,128 task instances across six scored sections and supports both text-only and multimodal settings, with 1,455 medical images disclosed according to their corresponding clinical stages. Appendix~\ref{app:experimental_details} provides the benchmark characteristics, data formatting, and implementation details required to interpret the experiments in this work.

\section{Method}
\label{sec:method}

\subsection{Overview}
\label{sec:method_overview}

\begin{figure*}[t]
  \includegraphics[width=\textwidth]{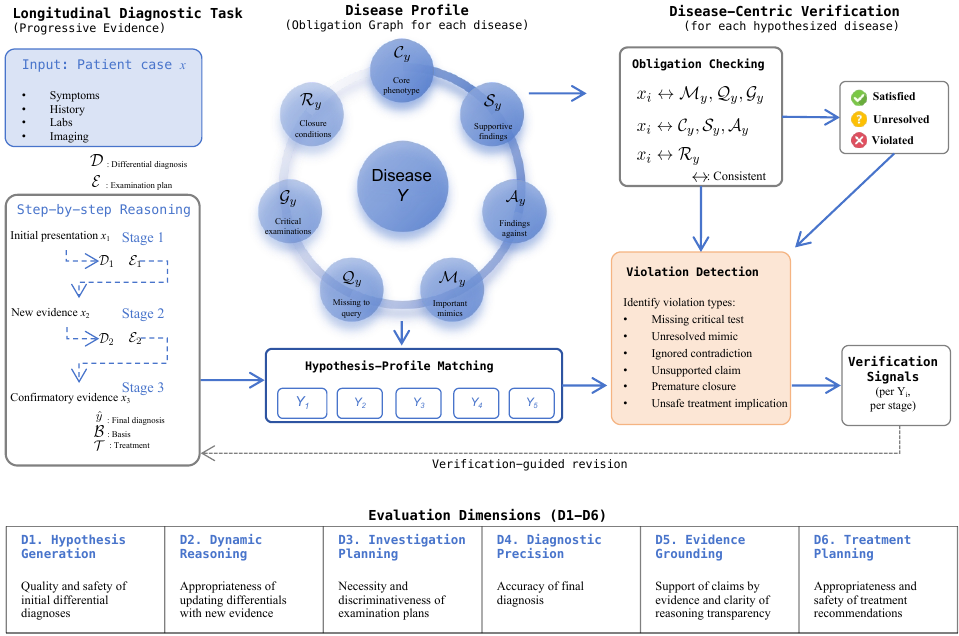}
  \caption{Overview of VeriDx. A longitudinal diagnostic trajectory is first generated under progressive evidence disclosure, including stage-wise differential diagnoses, examination plans, final diagnosis, diagnostic basis, and treatment plan. Each hypothesized disease is matched to a disease profile that encodes its clinical obligations, including phenotype consistency, supporting and opposing evidence, important mimics, critical examinations, missing information, and closure conditions. VeriDx checks these obligations for each hypothesis, labels them as satisfied, unresolved, or violated, and produces verification signals for detecting reasoning failures, guiding bounded revision, and evaluating trajectory-level clinical reasoning across D1--D6.
}
  \label{fig:illstation}
\end{figure*}

VeriDx is a disease-centric verification framework for longitudinal diagnostic reasoning. Its central idea is that a disease hypothesis creates clinical obligations: once a model proposes a diagnosis, it should check the evidence required by that disease, distinguish important mimics, resolve contradictions, request discriminative examinations, and justify when diagnostic closure is appropriate. This view is consistent with classical accounts of clinical reasoning as a hypothetico-deductive process of hypothesis generation, testing, revision, and closure \cite{eva_what_2005,norman_research_2005,cox_educational_2006}. It is also motivated by studies of diagnostic error, which show that unsafe diagnosis often results from incomplete data gathering, weak differential diagnosis, ignored evidence, and premature closure rather than from lack of medical knowledge alone \cite{croskerry_achieving_2002,graber_diagnostic_2005,schiff_diagnostic_2009,committee_on_diagnostic_error_in_health_care_improving_2015}.

Given a staged clinical trajectory, VeriDx links each model-generated disease hypothesis to a structured disease profile and verifies whether the corresponding obligations are satisfied, unresolved, or violated. The framework therefore shifts evaluation from asking only whether the final diagnosis is correct to asking whether the model has earned the right to make that diagnosis through a clinically responsible reasoning process.

\subsection{Longitudinal Diagnostic Task}
\label{sec:task_formulation}

We formulate diagnosis as a sequential decision-making task. For each case, the model observes a staged trajectory
\[
X = (x_1, x_2, x_3),
\]
where $x_1$ contains the initial presentation, $x_2$ introduces newly available examination results, and $x_3$ provides later confirmatory evidence. Information is disclosed progressively, so the model can only use evidence available up to the current stage.

The model produces:
\[
(\mathcal{D}_1, \mathcal{E}_1, \mathcal{D}_2, \mathcal{E}_2, \hat{y}, \mathcal{B}, \mathcal{T}),
\]
where $\mathcal{D}_1$ and $\mathcal{D}_2$ are the initial and updated differential diagnoses, $\mathcal{E}_1$ and $\mathcal{E}_2$ are the corresponding examination plans, $\hat{y}$ is the final diagnosis, $\mathcal{B}$ is the diagnostic basis, and $\mathcal{T}$ is the treatment plan. The expert reference trajectory is annotated as
\[
Y^\star =
(\mathcal{D}_1^\star, \mathcal{E}_1^\star, \mathcal{D}_2^\star, \mathcal{E}_2^\star, y^\star, \mathcal{B}^\star, \mathcal{T}^\star).
\]
Reference annotations distinguish required items from optional but acceptable items, allowing us to evaluate both critical coverage and over-generation.

\subsection{Disease Profiles}
\label{sec:disease_profiles}

For each disease $y$, VeriDx defines a disease profile:
\[
P_y =
(\mathcal{C}_y, \mathcal{S}_y, \mathcal{A}_y, \mathcal{M}_y, \mathcal{Q}_y, \mathcal{G}_y, \mathcal{R}_y).
\]
Here, $\mathcal{C}_y$ denotes the core phenotype, $\mathcal{S}_y$ supportive findings, $\mathcal{A}_y$ findings against the disease, $\mathcal{M}_y$ important mimics, $\mathcal{Q}_y$ missing information to check, $\mathcal{G}_y$ critical diagnostic examinations, and $\mathcal{R}_y$ closure conditions and disease-specific reasoning notes.

A disease profile is not an encyclopedia entry or a rigid diagnostic rule. Instead, it represents the obligations activated by a disease hypothesis. These obligations specify what evidence should support the disease, what evidence should weaken it, which alternatives must be distinguished, which examinations are necessary, and when diagnostic closure is justified. For imaging evidence, disease profiles encode guideline-derived imaging findings and required imaging examinations as clinically interpretable obligations. VeriDx checks these obligations against the clinical evidence extracted from multimodal model outputs. In this sense, profiles provide an auditable clinical structure while preserving the flexibility needed for incomplete and evolving real-world cases. This design follows the broader tradition of using explicit clinical knowledge structures to improve interpretability and safety in medical reasoning systems \cite{shortliffe_rule-based_1974,peleg_glif3_2000,scott_modelling_2023}.

Disease profiles encode default clinical obligations rather than deterministic decisions. Their application is conditioned on the evidence available for the current patient at each stage, including the initial presentation, exposure history, comorbidities, newly available examination results, and later confirmatory evidence. An obligation may therefore be satisfied, unresolved, or violated depending on the patient-specific context, allowing the same disease profile to produce different verification outcomes across patients. This design combines structured clinical knowledge with flexible patient-specific reasoning rather than applying guideline-derived obligations as fixed rules.


\subsection{Disease-Centric Verification}
\label{sec:verification}

Given a model-generated trajectory, VeriDx first extracts disease hypotheses, evidence claims, examination plans, exclusion statements, closure decisions, and treatment recommendations from the model output. Each disease hypothesis is then matched to its corresponding profile using exact, alias-based, and fuzzy disease-name matching.

For each matched hypothesis-profile pair, the verifier checks whether the current case evidence satisfies the profile-defined obligations. It asks whether the core phenotype is supported, contradicted, or unresolved; whether required evidence is present; whether findings against the disease are ignored; whether important mimics remain unresolved; whether critical tests are missing; and whether closure conditions have been met. Importantly, absence of evidence is not treated as evidence of absence. Unobserved but clinically relevant information is marked as missing or unresolved rather than negative, which is essential under progressive evidence disclosure.

VeriDx labels each obligation as satisfied, unresolved, or violated. This process converts broad reasoning defects into clinically interpretable error types. We focus on six common violation patterns: missing critical tests, unresolved mimics, ignored contradictions, unsupported claims, premature closure, and unsafe treatment implications. These errors correspond to well-studied sources of diagnostic failure, including insufficient data gathering, faulty synthesis, failure to consider alternatives, and premature closure \cite{croskerry_achieving_2002,graber_diagnostic_2005,schiff_diagnostic_2009,committee_on_diagnostic_error_in_health_care_improving_2015}.

When used for reasoning improvement, VeriDx performs bounded verification-guided revision. It may remove a diagnosis only when explicit counter-evidence is present, add an important missing mimic, supplement examinations that address workup gaps, or delay final closure when required obligations remain unresolved. The rewrite is constrained to local modifications rather than unrestricted regeneration, preserving acceptable parts of the original model output.

\subsection{Evaluation}
\label{sec:evaluation}

We evaluate each reasoning trajectory across six dimensions. \textbf{D1} measures initial hypothesis generation under information scarcity. \textbf{D2} measures dynamic diagnostic revision after new evidence appears. \textbf{D3} measures investigation planning, including whether required and discriminative tests are recommended. \textbf{D4} measures final diagnostic precision. \textbf{D5} measures evidence grounding and reasoning transparency. \textbf{D6} measures treatment planning quality and safety.

Together, D1--D6 evaluate the full diagnostic workflow: hypothesis generation, hypothesis revision, evidence acquisition, final diagnosis, evidence-grounded justification, and downstream clinical action. This allows VeriDx to distinguish models that only guess the correct final answer from models that maintain clinically responsible reasoning throughout the trajectory.

In the main experiments, we compare complete-information prediction with stepwise clinical reasoning, and evaluate the full VeriDx framework against ablated variants without disease profiles, phenotype checking, differential-diagnosis checking, or decision-closure control. We also report safety events such as dangerous diagnostic misses, diagnostic hallucination, hallucinated diagnostic basis, over-investigation, harmful treatment, and any unsafe event. Full prompts, extraction procedures, verifier rewriting constraints, judge rubrics, aggregation rules, and implementation details are provided in Appendix~\ref{app:experimental_details}.

\section{Results}

\begin{table*}[t]
\centering
\caption{Performance under progressively increasing levels of expert-annotated intermediate reasoning. \textbf{LLM-only Reasoning} uses model-generated intermediate states throughout the stepwise pipeline. \textbf{Gold $\mathcal{D}_1$/$\mathcal{E}_1$} replaces the first-stage differential diagnosis $\mathcal{D}_1$ and examination plan $\mathcal{E}_1$ with specialist annotations while preserving model-generated reasoning in subsequent stages. \textbf{Gold $\mathcal{D}_1$/$\mathcal{E}_1$/$\mathcal{D}_2$/$\mathcal{E}_2$} replaces all intermediate reasoning states with specialist annotations, representing the oracle upper-bound setting.}
\label{tab:performance-gap}
\small
\setlength{\tabcolsep}{5.2pt}
\begin{tabular}{lccc|ccc|ccc}
\toprule
& \multicolumn{3}{c|}{\textbf{LLM-only Reasoning}}
& \multicolumn{3}{c|}{\textbf{{Gold $\mathcal{D}_1$/$\mathcal{E}_1$}}}
& \multicolumn{3}{c}{\textbf{{Gold $\mathcal{D}_1$/$\mathcal{E}_1$/$\mathcal{D}_2$/$\mathcal{E}_2$}}} \\
\cmidrule(lr){2-4}
\cmidrule(lr){5-7}
\cmidrule(lr){8-10}
\textbf{Model}
& \textbf{Acc.} & \textbf{Recall} & \textbf{F1}
& \textbf{Acc.} & \textbf{Recall} & \textbf{F1}
& \textbf{Acc.} & \textbf{Recall} & \textbf{F1} \\
\midrule
AntAngelMed
& 0.20 & 0.38 & 0.28
& 0.27 & 0.45 & 0.34
& 0.71 & 0.74 & 0.72 \\

Baichuan-M3
& 0.30 & 0.37 & 0.35
& 0.35 & 0.44 & 0.40
& 0.75 & 0.77 & 0.76 \\

Claude-Opus-4.6
& 0.29 & 0.47 & 0.33
& 0.35 & 0.55 & 0.42
& 0.79 & 0.82 & 0.80 \\

Claude-Opus-4.6 (+Image)
& 0.31 & 0.47 & 0.37
& 0.37 & 0.56 & 0.45
& 0.81 & 0.84 & 0.82 \\

DeepSeek-v4-Pro
& 0.27 & 0.40 & 0.32
& 0.33 & 0.48 & 0.39
& 0.78 & 0.80 & 0.79 \\

Gemini-3.1-Pro
& 0.38 & 0.54 & 0.47
& 0.44 & 0.62 & 0.54
& 0.84 & 0.86 & 0.85 \\

Gemini-3.1-Pro (+Image)
& 0.39 & 0.53 & 0.47
& 0.45 & 0.61 & 0.54
& 0.85 & 0.87 & 0.86 \\

GPT-5.4
& 0.31 & 0.64 & 0.46
& 0.37 & 0.71 & 0.53
& 0.86 & 0.88 & 0.87 \\

GPT-5.4 (+Image)
& 0.33 & 0.58 & 0.40
& 0.39 & 0.66 & 0.48
& 0.87 & 0.89 & 0.88 \\

GPT-5.5
& 0.30 & 0.64 & 0.39
& 0.36 & 0.72 & 0.49
& 0.88 & 0.90 & 0.89 \\

GPT-5.5 (+Image)
& 0.31 & 0.68 & 0.41
& 0.38 & \textbf{0.75} & 0.52
& \textbf{0.89} & \textbf{0.91} & \textbf{0.90} \\

Kimi-K2.5
& 0.30 & 0.46 & 0.35
& 0.36 & 0.54 & 0.42
& 0.80 & 0.82 & 0.81 \\

Kimi-K2.5 (+Image)
& 0.30 & 0.46 & 0.34
& 0.36 & 0.54 & 0.41
& 0.81 & 0.83 & 0.82 \\

Qwen2.5-14B
& \textbf{0.56} & 0.62 & \textbf{0.56}
& \textbf{0.61} & 0.69 & \textbf{0.62}
& 0.82 & 0.84 & 0.83 \\

Qwen2.5-32B
& 0.53 & 0.54 & 0.51
& 0.58 & 0.61 & 0.57
& 0.83 & 0.85 & 0.84 \\

Qwen2.5-72B
& 0.43 & 0.55 & 0.46
& 0.49 & 0.63 & 0.54
& 0.84 & 0.86 & 0.85 \\
\bottomrule
\end{tabular}
\end{table*}

\begin{figure*}[t]
\centering
\includegraphics[width=\textwidth]{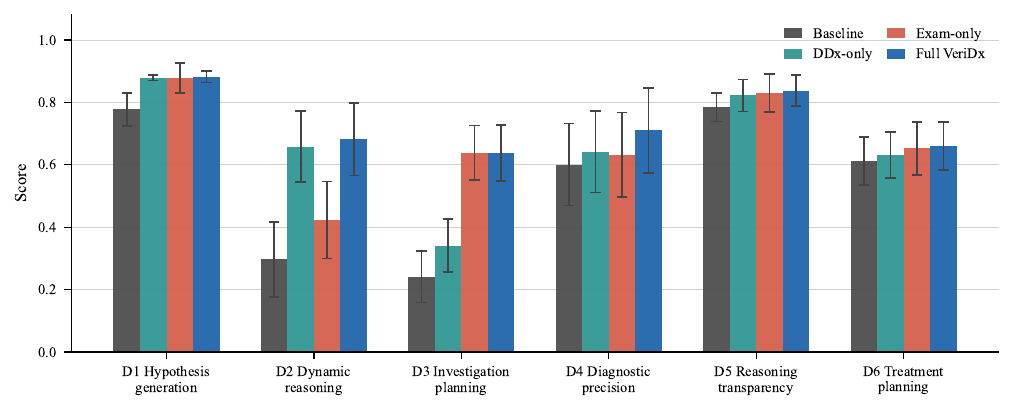}
\caption{
Task-level clinical reasoning performance of GPT-5.5 in the the multimodal setting across six evaluation dimensions and different verifier ablation settings. Bars indicate mean scores and error bars denote standard deviations across evaluation cases. DDx-only verification substantially improves dynamic reasoning and diagnostic precision, indicating that explicit differential-diagnosis correction mitigates premature diagnostic closure and promotes broader hypothesis exploration. In contrast, Exam-only verification primarily improves investigation planning, suggesting that structured examination completion enhances procedural completeness and diagnostic workup adequacy. The full VeriDx framework achieves the strongest overall performance across nearly all dimensions, demonstrating that differential-diagnosis verification and examination gap-filling provide complementary mechanisms for improving clinical reasoning safety, coherence, and diagnostic reliability.
}
\label{fig:results_base}
\end{figure*}


\subsection{Frontier LLMs Struggle with Sequential Clinical Reasoning}

As shown in Table~\ref{tab:performance-gap}, frontier multimodal LLMs achieve strong performance when complete clinical information is provided upfront. Under the full-information setting, most models can generate clinically plausible diagnoses and treatment recommendations, indicating that current systems already possess strong retrospective diagnostic capabilities.

However, their performance declines substantially under sequential clinical reasoning settings that more closely resemble real clinical workflows. When information is revealed progressively and models are required to iteratively update differential diagnoses, request examinations, and revise decisions over time, the quality of intermediate reasoning deteriorates markedly. Importantly, this degradation occurs even though each reasoning stage still contains sufficient information for appropriate clinical judgment.

The results further show that many models maintain relatively acceptable endpoint performance while exhibiting poor reasoning quality during the intermediate process. In practice, models frequently expand differential diagnoses excessively, fail to appropriately narrow hypotheses after receiving new evidence, and demonstrate unstable diagnostic refinement. 

\subsection{VeriDx Improves Clinical Reasoning}

As shown in Figure~\ref{fig:results_base}, we first evaluate the proposed verifier framework on GPT-5.5 across six dimensions of clinical reasoning. The baseline corresponds to GPT-5.5 under the multimodal setting, while the remaining conditions evaluate DDx-only verification, Exam-only verification, and the complete VeriDx framework. Relative to the baseline, DDx-only verification produces the most visible improvements in dynamic diagnostic revision and diagnostic precision. These results suggest that explicitly verifying the evolution of the differential diagnosis helps stabilize hypothesis updating and reduce premature diagnostic closure during sequential reasoning.

In contrast, Exam-only verification primarily improves investigation planning. By identifying missing examinations and promoting evidence acquisition before final decision-making, it produces a more complete diagnostic workup. However, its gains on higher-order reasoning dimensions remain comparatively limited. The complete VeriDx framework achieves the strongest overall performance across nearly all reasoning dimensions. Compared with either verification module alone, the combined framework simultaneously improves dynamic hypothesis refinement and procedural completeness, indicating that DDx verification and examination verification provide complementary benefits during clinical reasoning. Reasoning transparency, however, changes relatively little across the four settings.

To further assess whether these benefits extend across different model architectures, we apply the identical VeriDx pipeline to Gemini-3.1-Pro and Claude-Opus-4.6. As reported in Table~\ref{tab:cross_model_VeriDx} in Appendix~\ref{sec:cross_model_generalization}, the three evaluated model families exhibit a consistent overall pattern: DDx-only verification primarily improves dynamic diagnostic revision, whereas Exam-only verification produces its largest gains in investigation planning. Combining both modules achieves the strongest overall performance across most evaluation dimensions. These results provide evidence for the cross-model generalizability of VeriDx.



\subsection{Ablation Study}
\begin{table}[t]
\centering
\caption{Component ablation and profile-robustness results across six clinical reasoning dimensions for GPT-5.5 under the multimodal setting. The component-ablation settings use LLM-extracted DDx profiles. In the profile-robustness comparison, \textbf{VeriDx (LLM)} uses LLM-extracted DDx profiles, whereas \textbf{VeriDx (Gold)} uses specialist-constructed DDx profiles. All other experimental conditions are held fixed.}
\label{tab:ablation-robustness}
\small
\setlength{\tabcolsep}{4pt}
\begin{tabular}{lcccccc}
\toprule
\textbf{Setting}
& \textbf{D1}
& \textbf{D2}
& \textbf{D3}
& \textbf{D4}
& \textbf{D5}
& \textbf{D6} \\
\midrule
w/o profiles
& 0.80 & 0.36 & 0.33 & 0.62 & 0.79 & 0.48 \\
w/o phenotype
& 0.85 & 0.61 & 0.59 & 0.64 & 0.82 & 0.61 \\
w/o DDx check
& 0.86 & 0.65 & 0.60 & 0.69 & 0.74 & 0.56 \\
w/o closure
& 0.87 & 0.50 & 0.46 & 0.65 & 0.82 & 0.60 \\
\midrule
VeriDx (LLM)
& \textbf{0.88}
& 0.68
& 0.64
& 0.71
& 0.84
& 0.66 \\
VeriDx (Gold)
& \textbf{0.88}
& \textbf{0.71}
& \textbf{0.66}
& \textbf{0.72}
& \textbf{0.87}
& \textbf{0.68} \\
\bottomrule
\end{tabular}
\end{table}

\paragraph{Component ablation.}
As shown in Table~\ref{tab:ablation-robustness}, the ablation study further reveals that different VeriDx components contribute to distinct aspects of clinical reasoning performance. Removing DDx profiles leads to the largest overall performance degradation across multiple reasoning dimensions, particularly in hypothesis generation, dynamic reasoning, diagnostic precision, and treatment planning. The results suggest that DDx profiles provide the primary structural guidance for organizing and constraining the diagnostic hypothesis space. Removing phenotype checking mainly reduces diagnostic precision. The corresponding results indicate that phenotype consistency verification plays an important role in aligning predicted diagnoses with observed clinical manifestations during sequential reasoning. Removing DDx checking primarily affects reasoning transparency and treatment planning. Although its influence on overall diagnostic accuracy is comparatively smaller, the results suggest that this module improves explicit exclusion reasoning and promotes more interpretable diagnostic refinement. Finally, removing the closure check mainly weakens dynamic reasoning and investigation planning. Without closure control, models are more likely to terminate reasoning prematurely and perform insufficient evidence gathering before finalizing decisions.

\paragraph{Profile robustness.}
Because the DDx profiles used by VeriDx are automatically extracted from clinical guidelines, we assess their robustness by replacing them with gold-standard profiles constructed by specialist physicians using the same structured schema, while holding all other experimental conditions fixed. As shown in Table~\ref{tab:ablation-robustness}, VeriDx (Gold) yields modest improvements over VeriDx (LLM) across most dimensions, particularly in dynamic diagnostic revision (D2) and evidence grounding and reasoning transparency (D5), while initial hypothesis generation (D1) remains unchanged. These results suggest that specialist construction can further improve profile quality, while the automatically extracted profiles already preserve most of the clinical structure required for downstream verification. 

\section{Discussion}
Our findings expose an important limitation of current medical LLMs: strong performance in static, complete-information settings does not necessarily translate into reliable dynamic clinical reasoning. Although frontier models can map complete clinical vignettes to plausible final answers, their performance declines when they must navigate the sequential, incomplete, and evolving nature of clinical diagnosis. Without explicit structural guidance, models may fail to gather discriminative evidence, leave important mimics unresolved, or reach diagnostic closure before sufficient evidence has been obtained.

\paragraph{Redefining Diagnostic Evaluation.}
The primary implication of VeriDx is a shift in the evaluation of clinical reasoning from answer-centric accuracy toward disease-centric verification. By formalizing disease hypotheses as structured commitments, where each proposed diagnosis requires the model to verify phenotypes, resolve important mimics, and justify diagnostic closure, VeriDx provides a mechanism for auditing the intermediate reasoning process. Our ablation results show that disease profiles and verification mechanisms contribute to more stable hypothesis refinement and reduce the risk of premature closure. These findings suggest that diagnostic errors may arise not only from isolated hallucinations, but also from unresolved or violated reasoning commitments that propagate through the diagnostic trajectory.

\paragraph{A Neuro-Symbolic Bridge for Clinical AI.}
VeriDx connects the explicit structure of symbolic clinical knowledge with the flexibility of LLM-based reasoning. Symbolic systems provide interpretable rules and constraints but may struggle with the ambiguity, temporal evolution, and unstructured free text of real patient encounters. Conversely, LLMs can flexibly interpret evolving clinical narratives but may exhibit inconsistencies in evidence use and guideline adherence. VeriDx uses structured disease profiles as clinically grounded anchors while allowing the LLM to process longitudinal trajectories and extract patient-specific evidence. This design provides an explicit audit trail for disease-specific obligations while preserving the language-understanding capabilities required for complex clinical narratives.

\paragraph{Beyond Respiratory Medicine.}
Although the general framework of hypothesis-induced verification is not tied to a particular disease, its empirical generalizability to other specialties remains to be established. The profile structure may provide a template for other clinical domains. Beyond medicine, hypothesis-induced obligations may also offer a useful verification template for other high-stakes, multi-step reasoning tasks, although this possibility requires further empirical investigation.

\section{Limitations}
While VeriDx improves the structural integrity of LLM reasoning, several limitations remain. First, our framework currently relies on LLMs for both the extraction of intermediate claims and the downstream evaluation of whether obligations are met. Although tightly guided by structured profiles, this pipeline is still susceptible to cascading errors if the LLM incorrectly parses a complex clinical contradiction in the early reasoning steps. Second, the disease profiles require expert curation to ensure clinical validity. While we automated the initial generation from 2,563 guidelines, physician oversight remains necessary to ensure the clinical weight of specific mimics and required tests is appropriately calibrated. Third, VeriDx has currently been validated only in respiratory medicine, and its empirical generalizability to other specialties remains untested. Although the overall verification framework is disease-agnostic and potentially applicable across specialties, its extension would require newly curated longitudinal benchmarks, specialty-specific clinical guidelines, and expert annotation. Finally, our progressive disclosure evaluation strictly segments the diagnostic timeline into predefined steps. Real-world clinical workflows are often more continuous and asynchronous, requiring models to update hypotheses as individual lab results or imaging reports become available. Future work will extend VeriDx to support asynchronous evidence integration.


\clearpage

\bibliography{reference}

\appendix
\section{Ethical Considerations}

VeriDx is intended to support the safer clinical translation of medical LLMs by providing an auditable mechanism for verifying whether diagnostic reasoning satisfies disease-specific clinical obligations. Rather than replacing clinicians, the framework is designed to assist clinical evaluation, model auditing, and future decision-support development under qualified medical oversight. In this study, the disease profiles and benchmark annotations were reviewed by qualified clinicians to ensure that the encoded obligations, differential diagnoses, required examinations, and closure conditions were clinically meaningful. VeriDx was evaluated on a separately developed longitudinal pulmonary benchmark. All cases in this evaluation benchmark were de-identified before benchmark construction, and only information necessary for diagnostic reasoning evaluation was retained.

Despite these safeguards, VeriDx should not be used as an autonomous diagnostic system. Its verification results may still be affected by biases in the source guidelines, case selection, expert annotation, and LLM-based extraction or judgment. Therefore, its outputs should be interpreted as clinical decision-support signals rather than definitive medical conclusions. Such physician-guided, disease-centric verification may help identify unsafe reasoning patterns before deployment and promote more responsible clinical applications of medical LLMs. Future real-world use would require institutional review, privacy protection, bias auditing, prospective validation, and continuous monitoring in collaboration with clinical experts.




\section{Experimental Details}
\label{app:experimental_details}

\subsection{Data Processing}

Each case is stored as a structured JSON record with sections $S_1$--$S_9$. $S_1$ contains the initial patient presentation, $S_2$ the reference initial differential diagnosis, $S_3$ the reference initial examination plan, $S_4$ the first-stage examination results, $S_5$ the reference updated differential diagnosis, $S_6$ the reference follow-up examination plan, $S_7$ later key evidence, $S_8$ the final diagnosis and diagnostic basis, and $S_9$ the treatment plan and treatment response.

During loading, the textual content of each section is normalized by stripping empty fields and preserving clinically relevant image captions. If a section contains images, their local captions are appended to the section text as image-near-text descriptions. The raw image metadata, including local paths and URLs, is also preserved. In text-only experiments, image inputs are disabled. In text-image experiments, images are passed only when they belong to sections available at the current stage, preventing leakage from future evidence.

Reference outputs are extracted as follows. Initial and updated differential diagnoses are read from $S_2$ and $S_5$. Examination plans are read from the required and optional fields of $S_3$ and $S_6$. Final diagnosis and diagnostic basis are read from $S_8$, and treatment plan and treatment response from $S_9$. These fields are kept as lists so that open-ended model outputs can later be semantically evaluated.

\subsection{Closed-Loop Generation Protocol}

The implemented prediction pipeline follows a closed-loop progressive disclosure protocol. The model first receives only the initial patient presentation and generates an initial differential diagnosis and an initial examination plan:
\[
(R_2,R_3)=f(R_1).
\]
The system then extracts from $S_4$ only those examination results corresponding to tests ordered in $R_3$:
\[
R_4=g(S_4,R_3).
\]
The model receives $R_1$, $R_2$, $R_3$, and $R_4$ to generate an updated differential diagnosis and follow-up examination plan:
\[
(R_5,R_6)=f(R_1,R_2,R_3,R_4).
\]
Next, the system extracts from $S_7$ only key evidence relevant to the current differential diagnosis and ordered examinations:
\[
R_7=g(S_7,R_5,R_3,R_6).
\]
Finally, the model generates the final diagnosis and diagnostic basis, as well as the treatment plan and treatment response:
\[
(R_8,R_9)=f(R_1,R_2,R_3,R_4,R_5,R_6,R_7).
\]
This setting uses model-generated history by default, so errors in early differential diagnosis or examination planning can affect later available evidence and downstream decisions. Because the treatment response records in the original medical records are too brief, the generated treatment outcomes are for reference only and are not intended for evaluation purposes.

\subsection{Leakage Control}

Before Stage 1 generation, the initial presentation is checked for explicit final-diagnosis leakage. The model is asked whether $S_1$ directly states the confirmed diagnosis. If a leaked phrase is found, it is removed from the presentation and the cleaned text becomes $R_1$. This step prevents the model from solving the task by copying an explicitly stated final label.

\subsection{Generation Prompts}

All generation calls use the same system instruction:
\begin{quote}
You are a senior clinical physician. Output strictly valid JSON only---no extra text, no markdown fences. Never invent patient information not present in the provided context.
\end{quote}

For leakage checking, the model receives:
\begin{quote}
Determine whether the patient presentation contains a final-diagnosis leak. If no leak, output \texttt{\{"has\_leakage": false\}}. If leaked, output \texttt{\{"has\_leakage": true, "leaked\_phrase": "..."\}}.
\end{quote}

For Stage 1, the model is prompted to provide an initial assessment based only on $R_1$. It outputs:
\begin{verbatim}
{
  "differential_diagnoses": [
    {
      "name": "diagnosis name",
      "supporting_evidence": 
          ["brief phrases"],
      "opposing_evidence": 
          ["brief phrases"],
      "requirement": "required|optional"
    }
  ],
  "required": ["must-order tests"],
  "optional": ["optional tests"]
}
\end{verbatim}

For evidence extraction after Stage 1, the model is asked to extract from $S_4$ only results corresponding to the ordered tests in $R_3$:
\begin{quote}
Extract from $S_4$ only the results that correspond to the ordered tests in $R_3$. Ignore results for tests not listed in $R_3$. Output \texttt{\{"extracted\_text": "..."\}}.
\end{quote}

For Stage 2, the model receives $R_1$, prior DDx $R_2$, prior workup $R_3$, and extracted results $R_4$. It outputs:
\begin{verbatim}
{
  "differential_diagnoses": [
    {
      "name": "diagnosis name",
      "evidence": ["brief phrases"],
      "exclusion_reasons": 
          ["brief phrases"],
      "requirement": "required|optional"
    }
  ],
  "required": ["must-order tests"],
  "optional": ["optional tests"]
}
\end{verbatim}

For evidence extraction after Stage 2, the model is asked to extract from $S_7$ only findings directly relevant to the current DDx and ordered tests. For Stage 3, the model receives the accumulated trajectory and outputs:
\begin{verbatim}
{
  "final_diagnosis": ["diagnosis"],
  "diagnostic_basis": ["basis"],
  "treatment_plan": ["treatment item"],
  "treatment_response": ["response item"]
}
\end{verbatim}

\subsection{Disease-Profile Verification}

When verification is enabled, the generated DDx and examination plan are passed through a disease-profile module after Stage 1 and Stage 2. The profile library is loaded from the manually curated profile directory and the automatically generated profile directory. Each candidate diagnosis is resolved to a profile using exact, alias-based, and fuzzy disease-name matching. If no adequate profile exists, the system may generate a new profile for the missing disease and add it to the profile library.

Each disease profile specifies a disease-level diagnostic pathway, including core phenotype, supportive findings, findings against the diagnosis, poorly explained findings, important mimics, missing information to check, critical tests, and verifier notes. For each candidate diagnosis, the verifier compares the current case context and current examination plan with the matched profile. It records whether the core phenotype is supported, contradicted, or unresolved; whether supportive findings are present; whether exclusion findings are triggered; and whether critical tests are missing. Importantly, unresolved evidence is not treated as negative evidence.

The verifier performs bounded patch-style rewriting. A candidate diagnosis may be removed only when explicit counter-evidence is found in the profile comparison. The verifier may also add missing diagnoses suggested by the profile-guided assessment and add examinations that address profile-defined workup gaps. The rewrite is constrained to local modifications rather than unrestricted regeneration of the full answer.

\subsection{Verification Hints Across Stages}

Verifier outputs are propagated into later prompts as concise hints. Removed diagnoses are marked as excluded by strong counter-evidence and should not be re-added. Added diagnoses are marked as verifier-added candidates already included in the DDx. Added or replaced examinations are marked as verifier-added tests whose results may appear in later extracted evidence. This allows verification to affect the downstream closed-loop trajectory.

\subsection{Ablation Conditions}

The code evaluates three verification conditions. In the full setting, the verifier may modify both DDx and examinations. In the DDx-only setting, only DDx modifications are retained and the original examination plan is restored. In the Exam-only setting, only examination modifications are retained and the original DDx is restored. These ablations isolate whether observed changes are driven by hypothesis-space filtering or workup modification.

\subsection{LLM-as-Judge Evaluation}

Evaluation is performed with three structured judge calls per case. The judge uses clinical semantic matching rather than exact string matching, allowing equivalent medical expressions while penalizing vague, overly broad, unsupported, or hallucinated outputs.

Call A evaluates differential diagnosis quality. It scores the initial DDx for clinical appropriateness, comprehensiveness, whether key diagnoses are covered, and whether dangerous diagnoses are missed. If an updated DDx reference exists, it also evaluates whether the model updates its hypothesis space after new evidence, whether the primary diagnosis is present, whether the model converges appropriately, and whether negative evidence is used.

Call B evaluates examination planning. It scores whether required examinations are covered, whether the examination plan is clinically appropriate, whether completed tests are redundantly re-ordered, and whether the model over-investigates.

Call C evaluates final diagnosis, diagnostic basis, and treatment. It scores final diagnosis entity match, specificity, component completeness, and diagnostic hallucination. It also evaluates whether the diagnostic basis is factually accurate, complete, efficient, and free of hallucinated evidence. Treatment is evaluated for completeness, appropriateness, and harmful or contraindicated recommendations.

\subsection{Evaluation Dimensions}

The evaluator aggregates judge outputs into six workflow dimensions. D1 measures initial hypothesis generation under information scarcity. D2 measures dynamic diagnostic reasoning after new evidence. D3 measures investigation planning. D4 measures final diagnostic precision. D5 measures reasoning transparency and evidence grounding. D6 measures treatment planning quality and safety. Scores are normalized to a common scale for comparison across dimensions.

\subsection{Safety Metrics}

In addition to task scores, the evaluator records binary safety events: dangerous diagnostic miss, over-investigation, diagnostic hallucination, hallucinated diagnostic basis, harmful treatment, any hallucination, diagnosis-related unsafe event, and any unsafe event. For event type $e$, the reported event rate is:
\[
\mathrm{Rate}(e)
=
\frac{1}{N}\sum_{i=1}^{N} z_i(e),
\]
where $z_i(e)$ indicates whether event $e$ occurs in case $i$, and $N$ is the number of evaluated cases. Lower event rates indicate safer model behavior.

\subsection{Aggregation and Reporting}

For each run, the evaluator writes case-level score logs and an aggregate report. Cases that fail judge evaluation after repeated attempts are marked as missing and excluded from scored aggregates. Aggregate reports include the mean of each D1--D6 dimension and the rate of each safety event. Publication scripts then compare baseline, full verification, DDx-only, and Exam-only runs using the same evaluation pipeline.

\subsection{Cross-Model Generalization}
\label{sec:cross_model_generalization}

To determine whether the improvements shown in
Figure~\ref{fig:results_base} are specific to GPT-5.5, we extend the verification and rewriting experiments to Gemini-3.1-Pro and Claude-Opus-4.6. All three models are evaluated under the same multimodal setting using the identical VeriDx pipeline,
progressive-disclosure protocol, verification conditions, and
evaluation procedure. For each model, we compare the baseline
with DDx-only verification, Exam-only verification, and the
complete VeriDx framework.

\begin{table*}[t]
\centering
\small
\setlength{\tabcolsep}{3.5pt}
\caption{Cross-model verification and rewriting results across
six clinical reasoning dimensions. All models are evaluated
under the same multimodal setting using the identical VeriDx
pipeline. Values denote mean scores and standard deviations
across evaluation cases. Bold indicates the best result within
each model family.}
\label{tab:cross_model_VeriDx}
\resizebox{\textwidth}{!}{
\begin{tabular}{llcccccc}
\toprule
Model & Condition & D1 & D2 & D3 & D4 & D5 & D6 \\
\midrule

GPT-5.5 (+Image)
& Baseline
& $0.778 \pm 0.053$
& $0.297 \pm 0.120$
& $0.241 \pm 0.082$
& $0.601 \pm 0.131$
& $0.784 \pm 0.045$
& $0.613 \pm 0.077$ \\

& DDx-only
& $0.879 \pm 0.009$
& $0.658 \pm 0.114$
& $0.341 \pm 0.085$
& $0.642 \pm 0.131$
& $0.822 \pm 0.051$
& $0.632 \pm 0.073$ \\

& Exam-only
& $0.878 \pm 0.049$
& $0.423 \pm 0.123$
& $\mathbf{0.639 \pm 0.087}$
& $0.632 \pm 0.136$
& $0.830 \pm 0.061$
& $0.652 \pm 0.086$ \\

& Full VeriDx
& $\mathbf{0.882 \pm 0.018}$
& $\mathbf{0.682 \pm 0.116}$
& $0.638 \pm 0.089$
& $\mathbf{0.711 \pm 0.136}$
& $\mathbf{0.837 \pm 0.050}$
& $\mathbf{0.661 \pm 0.076}$ \\

\midrule

Gemini-3.1-Pro (+Image)
& Baseline
& $0.756 \pm 0.059$
& $0.354 \pm 0.120$
& $0.318 \pm 0.085$
& $0.594 \pm 0.134$
& $0.782 \pm 0.049$
& $0.618 \pm 0.080$ \\

& DDx-only
& $0.876 \pm 0.014$
& $0.661 \pm 0.115$
& $0.395 \pm 0.087$
& $0.641 \pm 0.131$
& $0.818 \pm 0.053$
& $0.631 \pm 0.074$ \\

& Exam-only
& $0.871 \pm 0.049$
& $0.482 \pm 0.122$
& $\mathbf{0.652 \pm 0.088}$
& $0.646 \pm 0.137$
& $\mathbf{0.836 \pm 0.060}$
& $\mathbf{0.667 \pm 0.084}$ \\

& Full VeriDx
& $\mathbf{0.882 \pm 0.018}$
& $\mathbf{0.678 \pm 0.117}$
& $0.646 \pm 0.090$
& $\mathbf{0.713 \pm 0.136}$
& $0.833 \pm 0.050$
& $0.663 \pm 0.077$ \\

\midrule

Claude-Opus-4.6 (+Image)
& Baseline
& $0.770 \pm 0.056$
& $0.332 \pm 0.124$
& $0.276 \pm 0.087$
& $0.602 \pm 0.137$
& $0.813 \pm 0.044$
& $0.629 \pm 0.074$ \\

& DDx-only
& $0.886 \pm 0.015$
& $0.632 \pm 0.119$
& $0.272 \pm 0.089$
& $0.646 \pm 0.135$
& $0.839 \pm 0.048$
& $0.640 \pm 0.073$ \\

& Exam-only
& $0.882 \pm 0.047$
& $0.449 \pm 0.127$
& $\mathbf{0.612 \pm 0.091}$
& $0.651 \pm 0.139$
& $\mathbf{0.846 \pm 0.058}$
& $0.659 \pm 0.084$ \\

& Full VeriDx
& $\mathbf{0.888 \pm 0.018}$
& $\mathbf{0.655 \pm 0.120}$
& $0.603 \pm 0.094$
& $\mathbf{0.701 \pm 0.137}$
& $0.842 \pm 0.051$
& $\mathbf{0.664 \pm 0.077}$ \\

\bottomrule
\end{tabular}
}
\end{table*}

As shown in Table~\ref{tab:cross_model_VeriDx}, a consistent
overall pattern is observed across the three model families.
DDx-only verification produces substantial improvements in
dynamic diagnostic revision (D2), while Exam-only verification
provides its largest gains in investigation planning (D3). The
complete VeriDx framework achieves the best performance on D1,
D2, and D4 for all three models and provides the strongest
treatment-planning performance for GPT-5.5 and Claude-Opus-4.6.
Although Exam-only verification slightly outperforms the complete
framework on several individual dimensions, particularly D3 and
D5, combining both modules yields the strongest overall
performance across most dimensions.

These findings indicate that the two verification modules retain
their complementary roles across GPT, Gemini, and Claude model
families. The observed improvements are therefore not restricted
to GPT-5.5, providing evidence for the cross-model generalizability
of VeriDx.

\section{Additional Details for Responsible Research}
\label{app:responsible_research}

\paragraph{Artifacts and intended use.}
This work uses and creates several scientific artifacts. Existing artifacts include clinical guidelines and reference documents used to construct disease profiles, as well as previously published medical LLMs, clinical reasoning benchmarks, and verification methods cited in the main paper. The primary artifact created in this work is the guideline-derived respiratory disease-profile library. VeriDx is evaluated on a separately developed longitudinal pulmonary benchmark containing de-identified real-world cases. The benchmark was developed independently of the VeriDx framework and is used in this work solely for evaluation. The disease-profile library is intended for research on clinical reasoning verification, model auditing, and decision-support evaluation, while the evaluation benchmark is used to assess longitudinal diagnostic reasoning under progressive evidence disclosure. Neither resource is intended to support autonomous diagnosis or replace qualified clinical judgment. Any use in clinical or educational settings should be conducted under institutional oversight and with qualified medical supervision. Where source documents or external tools have explicit access conditions, licenses, or terms of use, we follow the corresponding restrictions. Clinical cases are used only in de-identified form for research evaluation. Resources derived from clinical data are not intended for unrestricted public release unless permitted by the applicable data governance and ethics requirements.


\paragraph{Artifact documentation and dataset statistics.}
The guideline-derived corpus contains 2,563 respiratory guidelines and reference documents, occupying approximately 3.93 GB on disk. These documents are used to construct disease profiles covering a broad range of respiratory diseases. Each disease profile contains structured fields for core phenotype, supportive findings, findings against the disease, important mimics, missing information, critical diagnostic examinations, closure conditions, and treatment-relevant reasoning notes. The separately developed longitudinal pulmonary benchmark used for evaluation contains 188 diagnostically challenging real-world cases. Each case is organized into a standardized longitudinal trajectory with three diagnostic stages and nine structured sections: patient presentation, initial differential diagnosis, initial examination recommendation, examination results, refined differential diagnosis, further examination recommendation, key confirmatory evidence, final diagnosis and diagnostic basis, and treatment planning. Six sections are used as scored model outputs, yielding 1,128 task instances per model setting. The benchmark supports both text-only and multimodal evaluation. In the multimodal setting, 1,455 medical images are disclosed progressively according to the clinical stage at which they become available.


\paragraph{De-identification, privacy protection, and ethics review.}
The collection, de-identification, and research use of the clinical cases were reviewed and approved by the relevant institutional ethics review board under protocol CAMS\&PUMC-IEC-2024-067. All cases in the evaluation benchmark were de-identified before benchmark construction. Direct identifiers, including patient names, hospital identifiers, phone numbers, addresses, dates of birth, admission numbers, and other uniquely identifying administrative information, were removed. Dates and temporal descriptions were normalized or shifted when necessary while preserving clinically relevant temporal relationships, such as symptom duration, disease progression, examination timing, and treatment response. Only information necessary for diagnostic reasoning evaluation was retained, including clinically relevant symptoms, medical history, laboratory findings, imaging findings, pathology, microbiology, diagnostic decisions, treatment plans, and outcomes. Medical images were reviewed to ensure that any embedded identifiers were removed or masked before use. Because clinical narratives may contain rare disease combinations or unusual trajectories, de-identification was treated as an ongoing risk-control process rather than a one-time formatting step. We therefore combined automatic screening with manual clinical review to reduce the risk of retaining personally identifiable information.

\paragraph{Progressive evidence disclosure and leakage control.}
The separately developed longitudinal pulmonary benchmark is used in this work to evaluate diagnostic reasoning under incomplete and evolving information. Under its progressive evidence disclosure protocol, each case is divided into stages, and models receive only the evidence available up to the current clinical stage. Future information, including later confirmatory tests, treatment response, pathology, microbiology, and the final diagnosis, is withheld until the corresponding stage. For multimodal evaluation, medical images are disclosed only when they become available in the clinical trajectory, preventing models from using future imaging evidence to answer earlier diagnostic questions. During benchmark construction, textual summaries were checked to prevent accidental leakage of the final diagnosis or later-stage evidence into earlier sections. Each stage was also reviewed for explicit diagnosis names, retrospective statements, and confirmatory evidence that would not have been available at that point in the diagnostic process. Cases with unavoidable leakage were revised or excluded from the corresponding evaluation setting. In this work, we apply the established disclosure protocol without modifying the underlying benchmark structure.

\paragraph{Clinician review and human expert participation.}
Professional clinicians with expertise in respiratory medicine participated in the construction and review of the disease profiles. The separately developed evaluation benchmark and its annotations were also constructed and reviewed by qualified clinicians independently of the VeriDx framework. The human experts involved in this study were recruited through institutional and research collaborations rather than public crowdsourcing platforms. Their roles in this work included reviewing disease profiles, validating clinically meaningful obligations, assessing important differential diagnoses and discriminative examinations, performing annotation quality control, and refining the evaluation criteria. For each disease profile, clinicians reviewed whether the encoded phenotype, supportive evidence, opposing evidence, important mimics, critical examinations, treatment-relevant constraints, and closure conditions were clinically appropriate for verification. In the separately developed evaluation benchmark, clinicians reviewed the standardized longitudinal trajectory of each case and distinguished required items from optional but acceptable items, enabling evaluation of both critical coverage and over-generation. Clinicians were provided with structured review criteria specifying the expected scope of the disease profiles, the meaning of required versus optional diagnostic items, the criteria for identifying unresolved mimics and premature closure, and the scoring dimensions used in the evaluation. When disagreements arose during review, the relevant profile entries or benchmark annotations were discussed until a clinically acceptable consensus was reached. Ambiguous or insufficiently supported profile obligations were revised or removed. Compensation and participation arrangements followed the applicable institutional or collaboration agreements. No crowdworkers or non-expert public annotators were used for clinical judgment tasks.

\paragraph{Implementation and evaluation details.}
All models are evaluated under the same staged diagnostic protocol. For each case, the model receives progressively disclosed clinical evidence and is asked to generate outputs for the corresponding diagnostic stage, including differential diagnoses, examination recommendations, diagnostic bases, final diagnoses, and treatment planning. The same prompt templates and output schemas are used across models whenever possible. For closed-source API models, exact parameter counts, training data, and internal infrastructure are not publicly available. We therefore report model names, model versions when available, input settings, evaluation settings, and the number of task instances. No task-specific fine-tuning or hyperparameter search is performed. For open-source models, we report model identifiers and use the recommended inference settings unless otherwise specified. The VeriDx pipeline includes disease-name normalization, hypothesis extraction, evidence-claim extraction, profile matching, obligation checking, and result aggregation. Disease hypotheses are matched to profiles using exact matching, alias matching, and fuzzy name matching. Verification outputs are aggregated across six clinical reasoning dimensions: initial hypothesis generation, dynamic diagnostic revision, investigation planning, final diagnostic precision, evidence grounding, and treatment planning quality and safety. For statistical reporting, we aggregate scores across cases and model settings and report mean values. When applicable, we report standard deviations or error bars across evaluation cases. Safety events are counted at the case or trajectory level according to predefined error categories, including dangerous diagnostic misses, diagnostic hallucination, hallucinated diagnostic basis, over-investigation, harmful treatment, and any unsafe event.


\paragraph{Potential risks and safeguards.}
VeriDx is designed for verification, evaluation, and model auditing, not for autonomous diagnosis. Nevertheless, several risks remain. First, disease profiles may inherit biases, omissions, or differences in clinical practice patterns from the source guidelines and reference documents. Second, the separately developed evaluation benchmark may reflect the case mix, diagnostic practices, and documentation styles of the contributing clinical settings. Third, LLM-based extraction and judgment may introduce additional errors when parsing complex clinical narratives, negation, temporal evidence, or contradictory findings. To mitigate these risks, the disease profiles and benchmark annotations were reviewed by qualified clinicians, and verification outputs should be interpreted as decision-support signals rather than definitive medical conclusions. The framework is intended to identify unsafe reasoning patterns before deployment and to support the responsible clinical translation of medical LLMs under qualified medical oversight. Future real-world deployment would require prospective validation, institutional review, privacy protection, bias auditing, continuous monitoring, and integration with clinician-in-the-loop workflows.

\paragraph{Use of AI assistants.}
AI assistants were used to support language polishing, drafting assistance, and code development during the preparation of this work. They were not used as independent sources of scientific claims or clinical judgments. All scientific claims, experimental design choices, clinical content, annotations, results, and final manuscript text were reviewed and verified by the authors. Clinical content and disease-profile obligations were further reviewed by professional clinicians where relevant.

\end{document}